\newcommand{\CASKLoadStyle}{\usepackage[preprint,nonanonymous]{neurips_2026}}
\newcommand{\CASKAuthorEntry}[3]{%
  {\bfseries #1}\\
  #2\\
  \texttt{#3}%
}
\newcommand{\CASKAuthorOne}{%
  \CASKAuthorEntry{Buseong Kim}{d'strict Korea}{flight@skyline23.com}%
}
\newcommand{\CASKAuthorTwo}{%
  \CASKAuthorEntry{Heejun Gwon}{d'strict Korea}{contact@heejun.me}%
}
\newcommand{\CASKAuthorBlock}{%
  \CASKAuthorOne
  \and
  \CASKAuthorTwo
}
\newcommand{\CASKChecklistBlock}{}

\documentclass{article}

\CASKLoadStyle

\usepackage[utf8]{inputenc}
\usepackage[T1]{fontenc}
\usepackage{hyperref}
\usepackage{url}
\usepackage{booktabs}
\usepackage{longtable}
\usepackage{array}
\usepackage{amsfonts}
\usepackage{amsmath}
\usepackage{amssymb}
\usepackage{nicefrac}
\usepackage{microtype}
\usepackage{xcolor}
\usepackage{enumitem}
\usepackage{graphicx}
\usepackage{tabularx}
\usepackage{float}

\newcolumntype{Y}{>{\raggedright\arraybackslash}X}
\setlist[itemize]{leftmargin=1.5em}
\setlist[enumerate]{leftmargin=1.5em}
\graphicspath{{figures/}}

\title{CASK: Core-Aware Selective KV Compression for Reasoning Traces}
\author{\CASKAuthorBlock}
\begin{document}
\maketitle

\begin{abstract}
In large language models performing long-form reasoning, the KV cache grows rapidly with decode length, creating bottlenecks in both memory and stability. Existing reasoning-oriented KV compression has largely remained within an eviction-centered perspective that estimates per-token importance more precisely in order to discard less important tokens. However, our preliminary analysis showed that scorer refinement alone struggles to substantially reorganize the actual keep-set and therefore may not be the primary lever governing reasoning behavior. Building on this observation, this paper redefines reasoning KV compression not as a simple ranking problem but as a \textbf{behavior-preserving structured consolidation problem}. CASK partitions the decode-time reasoning trace into a \textbf{protected core} that directly contributes to answer formation and state anchoring, and \textbf{mergeable scratch} with high redundancy. The core is protected, and selective consolidation is applied only to the scratch. To address the failure mode in prompt-heavy regimes where the prefix pre-empts the budget and deactivates decode-stage compression, CASK further introduces \textbf{two-stage compression} that applies eviction to the prefix and selective consolidation to the decode stage. On the H100 reasoning gate, CASK shows higher full-KV continuation fidelity than TriAttention at the same budget on both AIME24 and AIME25, with recurring \textbf{cask@384 > triattention@512} crossings. In prompt-heavy replay, \textbf{multi\_news} and \textbf{vcsum} emerge as replay-level decode-active same-budget witnesses, while \textbf{qmsum} and \textbf{gov\_report} reveal the \textbf{prefix\_budget\_exhausted} boundary. Moreover, the actual-output layer is not summarized by a single lexical-overlap metric; instead, it is read separately through \textbf{official task metric}, \textbf{lexical overlap}, and \textbf{semantic/reference similarity}. The most defensible conclusion supported by current evidence is therefore that the key to reasoning KV compression lies not in more complex scorer engineering but in \textbf{combining core preservation with selective scratch consolidation to lower the usable budget frontier}.
\end{abstract}

\section{Introduction}

LLMs performing long-form reasoning generate lengthy decode traces involving intermediate derivation, backtracking, self-verification, and alternative branch exploration in order to produce a single correct answer. In this process, each decode step accesses the key-value states of past tokens, so the KV cache grows proportionally to sequence length. KV compression is therefore essential for reducing both reasoning latency and memory usage. However, reasoning workloads present structural challenges distinct from simple long-context understanding, because within the reasoning trace, states directly connected to answer formation coexist with exploratory scratch work used to reach those states.

Existing reasoning-oriented KV compression has generally treated this as a problem of more precise importance ranking: assign scalar importance to each token, retain the top tokens, and discard the rest. This perspective is natural for retrieval-oriented long-context settings or for general decode pruning. However, in reasoning traces, not all non-core tokens are equal. Some tokens, if removed, immediately cause state anchoring to collapse, while others may not require individual preservation as long as a sufficiently similar representative remains. In other words, reasoning traces exhibit role differentiation rather than a simple importance ordering. Recent reasoning-oriented compression methods such as TriAttention and R-KV improved the scoring and eviction side of the problem, but they still remain fundamentally within an eviction-centered framing \citep{mao2026triattention,cai2025rkv}.

The starting point of this work is precisely this observation of role differentiation, and more specifically, a reflection on \textbf{the limitations of Phase 1}. In Phase 1, we approached the problem through scorer-side refinements such as adaptive horizon, RMS2, and variational horizon. That process was still useful. We obtained mathematical assets including the horizon kernel language, score-family interpretations, and frequency-aware diagnostics, and gained a clearer understanding of what kinds of deviations can be captured at the ranking level. However, a more important fact also emerged: unlike the mathematical differences in kernel space, the actual keep-set churn was limited, and the set of protected tokens did not substantially reorganize even when the scorer changed.

This observation directly motivated the pivot. If scorer refinement cannot substantially change the actual keep-set, then the key lever for reasoning behavior may not lie in more sophisticated scalar ranking itself. In other words, the problem is not only about better predicting "which token is more important"; it may instead lie in having \textbf{incorrectly defined what constitutes the preservation target versus the consolidatable target} in the first place. CASK's current core-aware selective consolidation perspective arose directly from this failure. Phase 1 is not a discarded predecessor but an analytical foundation that supplied the rationale for why-not-scorer and enabled the transition to the current policy design. \textbf{The motivation, goals, failure points, and pivot rationale of Phase 1 are detailed in Appendix A; the mathematical assets of Phase 1 are in Appendix B.}

This paper re-poses reasoning KV compression as follows: rather than treating all non-core tokens as candidates for discard, can we first determine what must be preserved and what can be safely merged? To answer this question, we propose CASK, Core-Aware Selective KV Compression for Reasoning Traces. CASK decomposes the decode-time trace into protected core and mergeable scratch, preserves the core, and performs selective consolidation only on the scratch. Additionally, in prompt-heavy regimes, it introduces a two-stage design combining prefix eviction and decode consolidation to prevent method coverage itself from being blocked.

The contributions supported by current evidence can be conservatively summarized as follows.

\begin{enumerate}
\item We propose a core-aware selective consolidation perspective that decomposes the reasoning trace into protected core and mergeable scratch.
\item We propose two-stage compression combining prefix eviction and decode consolidation to handle prompt-heavy regimes.
\item We present a bounded empirical interpretation that jointly reads representative-set mass, regime guard, and decode activity.
\item We organize the H100 reasoning gate, prompt-heavy replay, and actual-output bridge into a single claim-disciplined empirical package.
\end{enumerate}

Here, teacher-forced replay should be positioned not as an independent contribution but as an evaluation protocol required at the current stage. This distinction matters. The center of this paper lies not in proposing a new evaluation framework but in the structural argument that preservation and consolidation must be separated within reasoning traces.

The headline directly supported by current results should also align with this contribution structure. CASK's strength does not lie in compressing most aggressively across all settings. Rather, it lies in lowering the minimum usable budget required to maintain the same full-KV behavior. Accordingly, this paper should position CASK not as a more aggressive compressor but as a behavior-preserving policy that shifts the usable budget frontier downward.

\section{Method}

\subsection{Core-aware decomposition}

CASK's starting point is the observation that tokens within the reasoning trace are not equally important. Some tokens directly anchor the state of subsequent reasoning, while some are closer to scratch generated during intermediate exploration that repeat similar functions. Without reflecting this difference and treating the entire trace as a single ranked list, states that should not be merged and states where only a representative needs to remain are processed under the same primitive.

To address this, CASK decomposes the active decode trace into two parts. The first is the \textbf{protected core}, and the second is the \textbf{mergeable scratch}. The protected core includes states directly connected to answer formation, intermediate state anchoring, and recent reasoning pivots. This set is not sent to the merge target even when subject to compression. Conversely, the mergeable scratch consists of tokens with high redundancy such as repeated exploration, restatements, self-checks, and near-duplicate derivations, and CASK performs representative construction only on this part.

The key point is that this decomposition is not merely an interpretive description but an actual policy. CASK first determines what to preserve, and only then decides what to consolidate. That is, the center of compression lies not in a single discard rule but in the policy structure itself that separates preserve and consolidate.

\textbf{Figure 5} summarizes this method-level structure. Protected core is preserved, selective consolidation is applied only to mergeable scratch, and the two-stage design explicitly separates prefix-side slack creation from decode-side structured consolidation.

\begin{figure}[t]
\centering
\includegraphics[width=\linewidth]{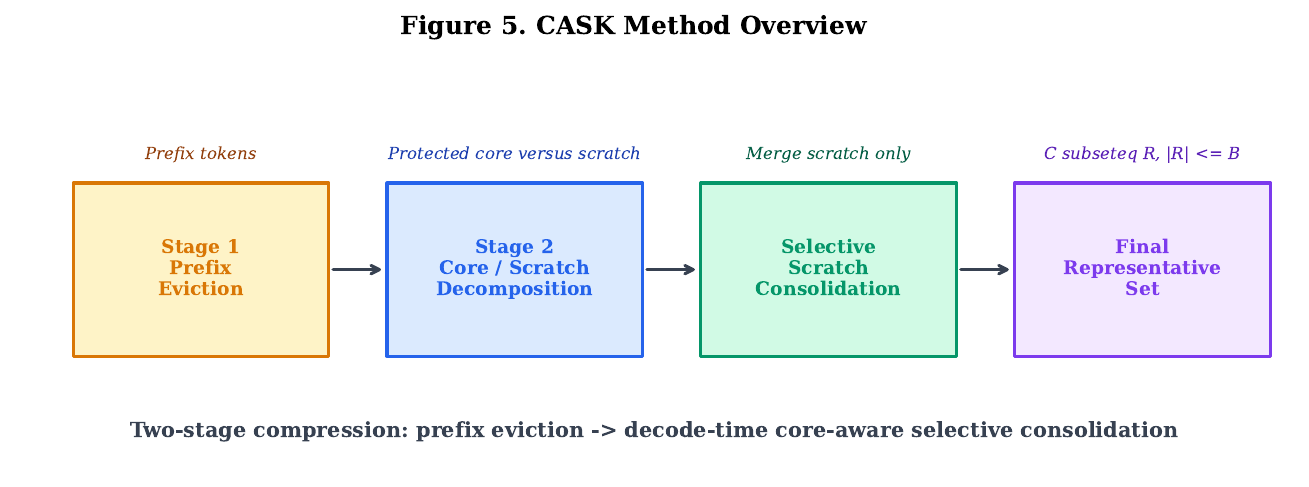}
\caption{Method overview. CASK separates prefix-stage eviction from decode-stage selective consolidation, while enforcing protected core preservation inside the decode trace.}
\end{figure}

\subsection{Two-stage compression}

In prompt-heavy regimes, another problem arises. When the prefix occupies nearly the entire budget, the decode-stage compression path may not open at all. In this case, before discussing the quality of the method, method coverage itself collapses. Even if a decode-stage merge is designed, the path may remain deactivated throughout the actual run.

CASK addresses this with \textbf{two-stage compression}. In Stage 1, eviction is applied to the prefix to secure slack for the decode stage. The goal is not to best compress the prefix but to leave budget space for decode-stage consolidation to actually activate. In Stage 2, protected core preservation and selective scratch consolidation are applied to the decode trace. This design is based on the observation that prefix and decode are structurally different objects. The prefix is given context, while the decode is the reasoning workspace the model dynamically uses. There is no reason to treat both intervals with the same compression primitive.

The significance of this two-stage design is not merely engineering supplementation. In prompt-heavy settings, the core narrative of CASK is not just the decode-stage merge itself but restructuring the policy so that the prefix does not block decode-stage method coverage. Accordingly, witnesses like \textbf{qasper} should be read more accurately as two-stage coverage and crossing evidence rather than as decode-stage superiority evidence.

\subsection{Representative-set objective}

Conceptually, CASK's goal is to construct a representative set that maximally preserves reference full-KV behavior under budget constraints. This can be written as follows.

$$
\max_{R} \; \mathrm{Fidelity}(R; \mathrm{FullKV})
\quad
\text{s.t.}
\quad
|R| \le B,
\; C \subseteq R
$$

Here, $C$ is the protected core, $R$ is the final representative set, and $B$ is the available physical budget. The important constraint in this expression is $C \subseteq R$: the protected core must be included in the final representative set. Scratch is consolidated into representatives only within the remaining budget. This formulation clearly shows that CASK is not a method that learns a better ranking function but one that first reflects role differentiation within the reasoning trace.

\subsection{Basic m-folding construction}

CASK's local merge can be understood not as simply fixing one token as a representative but as folding information within the merge group to create a representative. For a merge group $G$ with local mass $m_G$, the most basic folding form can be written as:

$$
m_G = \sum_{i \in G} a_i,
\qquad
\tilde{k}_G = \frac{1}{m_G} \sum_{i \in G} a_i k_i,
\qquad
\tilde{v}_G = \frac{1}{m_G} \sum_{i \in G} a_i v_i
$$

Here, $a_i$ is a non-negative weight reflecting local score mass, similarity weight, or position-aware importance within the group. This formula reads CASK's merge not as "a process of arbitrarily leaving one representative token" but as \textbf{a process of folding representative key/value while preserving group mass}. In actual implementation, the definition of $a_i$ may vary, but the key point needed at the main text level is that CASK treats merge as representative construction rather than simple discard.

\subsection{Kappa-weighted merge geometry}

CASK's local merge is not interpreted solely through raw cosine similarity. Using the horizon-kernel language obtained from Phase 1, it can be read as a geometry that weighs differences in frequency bands with greater future relevance more heavily. The most basic kernel definition is:

$$
\kappa_{\pi_h}(\omega)
:=
\mathbb{E}_{\delta \sim \pi_h}
\left[e^{i\omega\delta}\right]
$$

Using this, the local merge distance between two scratch tokens $i, j$ can be written as:

$$
d_{\kappa}(k_i, k_j)
=
\sum_f
\left|\kappa_{\pi_h}(\omega_f)\right|
\cdot
\|k_{i,f} - k_{j,f}\|_2
$$

The meaning of this formula is clear. Differences in bands with almost no future relevance matter less for merge decisions, while differences in bands that contribute more to actual future query behavior should be reflected more strictly. Thus, CASK's merge is more accurately interpreted as \textbf{future-relevance-aware local geometry} rather than simple proximity.

\subsection{Representative-set mass diagnostics}

Judging compression quality solely by saved tokens makes it easy to miss representative collapse. For that reason, the main text also examines how much oracle-relevant mass the protected core and final representative set preserve.

$$
\rho_{\mathrm{core}}
=
\frac{\sum_{k \in C} s_k^{\mathrm{oracle}}}{\sum_{k \in \mathrm{TopK}} s_k^{\mathrm{oracle}}},
\qquad
\rho_{\mathrm{rep}}
=
\frac{\sum_{k \in R} s_k^{\mathrm{oracle}}}{\sum_{k \in \mathrm{TopK}} s_k^{\mathrm{oracle}}}
$$

With an approximate perturbation bound, representative quality can be read as the sum of within-group dispersion and lost mass.

$$
\left|\sum_{i \in G} a_i \langle q, k_i \rangle - \langle q, \tilde{k}_G \rangle\right|
\lesssim
\|q\|_{\kappa,*} \sum_{i \in G} a_i \|k_i - \tilde{k}_G\|_{\kappa}
+
|\Delta m_G|
$$

The first term reads as within-group $\kappa$-dispersion, and the second as lost mass. In the main text, this formula is used not as a theorem but as diagnostic intuition; the full formula archive is in Appendix B.

Here, $s_k^{\mathrm{oracle}}$ denotes oracle top-k score mass. $\rho_{\mathrm{core}}$ represents the mass directly protected by core alone, and $\rho_{\mathrm{rep}}$ represents the total mass recovered by the final representative set. In healthy regimes, $\rho_{\mathrm{rep}}$ remains high, while in collapse regimes, the representative set itself breaks down and this value drops sharply.

\subsection{Why this is not just a better scorer}

Reading CASK as a better scorer paper misses the point. The actual narrative lies in role separation. Within reasoning traces, states that must not be merged and states where a good representative suffices coexist. Scorer refinement can help better estimate the relative importance of individual tokens, but it does not automatically resolve which regions should be entirely excluded from merge targets.

CASK's central claim is therefore as follows. The key lever of reasoning KV compression lies not in designing more complex scalar ranking but in \textbf{structurally separating what to preserve from what to consolidate}. This difference should be interpreted as the cause of same-budget fidelity gains and usable-budget frontier shifts.

\subsection{Method-level claim boundary}

Based on current evidence, claims must be conservatively bounded even in the method section. CASK does not aim to be the most aggressive compressor across all settings. Nor can it claim that decode-stage merge is front-and-center across all witnesses. A more accurate statement is that CASK reflects role differentiation in reasoning traces at the policy level, secures method coverage through a two-stage policy in prompt-heavy regimes, and consequently forms a stronger budget-fidelity frontier in some witnesses.

That is, the method's core is \textbf{behavior-preserving policy design}, not universal superiority. Maintaining this framing prevents results and limitations from conflicting.

\section{Experimental Protocol}

For prompt-heavy and bridge analysis, we use tasks drawn from the LongBench evaluation suite \citep{bai2024longbench}.

\subsection{Evaluation principle}

At the current stage, CASK's primary evaluation metric is not sparse exact-match accuracy but \textbf{teacher-forced reference fidelity}. In reasoning benchmarks, single-run accuracy can become excessively sparse due to sample variance, truncation, answer formatting, and extraction noise. By contrast, teacher-forced replay directly observes when and how closely the compressed policy follows the reference full-KV continuation. In the current manuscript, replay is therefore used not as an independent contribution but as an evaluation protocol for reading behavior preservation without exaggeration.

This principle connects to the paper's headline. If CASK's goal is "lowering the minimum usable budget required to maintain the same behavior" rather than "discarding more at the same budget," then metrics that more directly measure the distance from the reference continuation should take priority over sparse metrics that only examine a single final answer. At the same time, when reading the actual-output bridge, conclusions are not drawn from a single lexical-overlap score. Output-level evaluation instead reads along three axes: \textbf{official task metric}, \textbf{lexical overlap}, and \textbf{semantic/reference similarity}, where the semantic axis serves as an auxiliary metric measuring how semantically close the output is to the reference continuation.

\subsection{Teacher-forced replay procedure}

The evaluation procedure is as follows. First, the prompt and continuation are obtained from a full-KV or trusted high-budget reference run. Then the same prompt is fed to the candidate compression method, and the reference continuation is replayed token by token via teacher forcing. At each step, the candidate's next-token distribution and runtime KV statistics are recorded. This procedure directly measures exactly when behavior drift due to compression begins and how rapidly probability mass collapses after divergence.

The advantage of this protocol is that it allows decomposing the structure of drift. Some methods may maintain probability mass relatively stably after the first mismatch even if it comes early, while others may see NLL deteriorate much faster after diverging at the same point. Replay is therefore far more informative than a simple "correct/incorrect" binary.

\subsection{Primary metrics}

The manuscript's central metrics are four. The first is \textbf{top-1 agreement}, the second is \textbf{top-5 coverage}, the third is \textbf{mean NLL}, and the fourth is \textbf{first mismatch}. Terminal saved ratio is used only as an auxiliary metric showing compression aggressiveness.

Let $T$ be the replay length, $p_t$ be the candidate next-token distribution at time $t$, and $y_t^{\mathrm{ref}}$ be the reference token. Each metric is defined as follows.

\textbf{Top-1 agreement}

$$
\mathrm{Top1}
=
\frac{1}{T}
\sum_{t=1}^{T}
\mathbf{1}\left[\arg\max p_t = y_t^{\mathrm{ref}}\right]
$$

\textbf{Top-5 coverage}

$$
\mathrm{Top5}
=
\frac{1}{T}
\sum_{t=1}^{T}
\mathbf{1}\left[y_t^{\mathrm{ref}} \in \mathrm{Top5}(p_t)\right]
$$

\textbf{Mean negative log-likelihood}

$$
\mathrm{NLL}
=
-
\frac{1}{T}
\sum_{t=1}^{T}
\log p_t\left(y_t^{\mathrm{ref}}\right)
$$

\textbf{First mismatch}

$$
t_{\mathrm{mis}}
=
\min \left\{ t : \arg\max p_t \neq y_t^{\mathrm{ref}} \right\}
$$

In the actual-output bridge, output-level proximity is read along three axes separately from the replay metrics above. The first axis is lexical overlap; in the current manuscript, we use the normalized sequence similarity between the reference continuation and candidate continuation.

$$
\mathrm{SeqRatio}
=
\frac{\mathrm{LCS}(\hat{y}_{1:L}, y^{\mathrm{ref}}_{1:L'})}{\max(L,L')}
$$

Here, $\hat{y}_{1:L}$ is the candidate output, $y^{\mathrm{ref}}_{1:L'}$ is the reference output, and $\mathrm{LCS}$ is the longest common subsequence length. The second axis is semantic/reference similarity, defined as the cosine similarity of representations obtained by a sentence encoder $\phi(\cdot)$.

$$
\mathrm{SemSim}
=
\frac{\langle \phi(\hat{y}), \phi(y^{\mathrm{ref}}) \rangle}{\|\phi(\hat{y})\|_2\,\|\phi(y^{\mathrm{ref}})\|_2}
$$

The third axis is the benchmark-native official task metric. Since this varies by task, it is not fixed into a single common formula; instead, the dataset-specific evaluator provided by LongBench is used as-is.

$$
\mathrm{TaskMetric}
=
\mathrm{Eval}_{\mathrm{task}}(\hat{y}, y^{\mathrm{ref}}, \mathcal{A})
$$

Here, $\mathcal{A}$ denotes the task-specific answer annotation or reference set. The reason for having these three axes together is that lexical overlap alone can undervalue cases like \textbf{vcsum} where expressions differ but the meaning is semantically closer to the reference.

These four metrics serve different roles. Top-1 agreement shows how often the reference continuation is exactly followed. Top-5 coverage shows whether the candidate still keeps the reference token within the plausible set. Mean NLL measures the probability mass assigned to the reference token continuously, providing a more sensitive fidelity signal than simple rank mismatch. First mismatch intuitively shows when divergence begins, but alone it can be non-monotonic, so it must be interpreted together with top-1 or mean NLL.

Meanwhile, the actual-output bridge does not use replay metrics directly as headlines. There, (1) LongBench's \textbf{official task metric}, (2) \textbf{lexical overlap} against the reference output, and (3) \textbf{semantic/reference similarity} against the reference output are examined together. Lexical overlap is sensitive to reference phrasing restoration, while semantic similarity better reflects whether the same summary or response was produced even if phrasing differs. To correctly interpret cases like \textbf{vcsum}, where replay fidelity is strong but lexical score is low while the actual content reads closer to the reference, all three axes must therefore be used together.

\subsection{Terminal saved ratio}

The main text's core metrics are top-1, top-5, mean NLL, and first mismatch, but terminal saved ratio is also recorded as an auxiliary metric for viewing compression aggressiveness.

$$
\mathrm{SavedRatio}_{\mathrm{term}}
=
1 - \frac{\text{terminal cache tokens}}{\text{terminal total cardinality}}
$$

This value literally shows how much the cache was reduced at the final time step. Still, this number does not substitute for fidelity.

\subsection{Why saved ratio is not the headline}

Saved ratio is important but not the headline metric. Even if some method removes more tokens at the same nominal budget, if it loses reference behavior faster as a consequence, it is hard to call it better from the perspective of reasoning compression. In fact, in current evidence, CASK is less aggressive or has similar saved ratio in some slices compared to TriAttention but maintains higher fidelity. What matters here is not "how much was discarded" but "how long reference behavior is maintained at that budget."

This paper therefore does not place fixed-budget savings at the center of its claims as the absolute advantage. Instead, it uses the reduction in the minimum usable budget required to maintain the same full-KV behavior as the core interpretation axis.

\subsection{Witness design and regime separation}

The current witness set is constructed by selecting high-signal cases that reveal different regimes, rather than collapsing into a single benchmark average. The short-prompt reasoning witness is \textbf{hexagon}, and for prompt-heavy settings: \textbf{qasper}, \textbf{multi\_news}, \textbf{hotpotqa}, \textbf{musique}, \textbf{2wikimqa}, plus follow-up probes \textbf{vcsum}, \textbf{qmsum}, \textbf{gov\_report}.

What matters here is not treating all witnesses as the same kind of evidence. \textbf{multi\_news} and \textbf{vcsum} are replay witnesses where decode-stage activity is actually observed. By contrast, \textbf{qasper} is strong crossing evidence but should be read as a prefix-dominant regime. \textbf{qmsum} and \textbf{gov\_report} are follow-up probes revealing the \textbf{prefix\_budget\_exhausted} boundary, not evidence for asserting the method's universal superiority but evidence that makes the claim boundary more rigorous. \textbf{2wikimqa} remains a retained boundary case.

That is, the core of the experimental protocol lies not simply in defining metrics. \textbf{The discipline of distinguishing what witness is evidence of what} is itself part of the current empirical package.

\subsection{Compute and implementation resources}

All headline replay and actual-output packages use \textbf{Qwen3-8B} loaded in \textbf{bfloat16} through the \textbf{sdpa} attention path. The main paper-facing packages are the tracked \textbf{H100 PCIe} runs in \texttt{artifacts/h100\_2026\_04\_10/} and \texttt{artifacts/h100\_2026\_04\_11/}. Separate local sanity checks and cheap witness debugging runs are explicitly labeled as local packages and were executed on a single \textbf{RTX 5070 Ti 16 GB} worker.

The paper does not collapse these heterogeneous jobs into a single misleading wall-clock headline. Instead, each package carries the concrete reproduction information needed for the reported rows: hardware class, model path, budget grid, command line, and raw output roots. In particular, \texttt{fidelity\_manifest.json}, \texttt{overnight\_manifest.json}, \texttt{actual\_bridge\_summary.json}, and \texttt{artifacts/COMMAND\_MAP.md} record the exact command patterns and file provenance for the tracked H100 reasoning gate, prompt-heavy replay package, and actual-output bridge. Practically, local replay debugging fits on a single 16 GB consumer GPU, while the headline reasoning and prompt-heavy packages were run on H100-class workers.

\subsection{Claim-disciplined reading of results}

The way results are read under this protocol must also be constrained. The H100 reasoning gate is the core evidence showing same-budget fidelity advantage and some budget crossings. The prompt-heavy replay package shows strong same-budget gains, but within it, decode-active witnesses, prefix-dominant witnesses, and boundary cases must be separated. The actual-output bridge is an auxiliary axis showing that replay-level gains are not entirely unrelated to actual generation.

The interpretation of results in this manuscript is therefore not "always better across all settings." A more accurate statement is that under the current protocol, CASK shows a stronger budget-fidelity frontier across multiple witnesses, and interpreting this structure without exaggeration requires jointly reading replay fidelity, regime guard, and decode activity. \textbf{Raw replay counts and raw audit tables are in Appendix C; witness/package role assignments are in Appendix D.}

\section{Results}

This section restructures results around three questions. First, does the same-budget fidelity advantage recur in H100 reasoning slices. Second, what kind of witnesses does two-stage CASK actually produce in prompt-heavy regimes. Third, does replay-level fidelity gain connect to actual generation. The purpose of this section is not to relist all raw tables but to build claims stably by separating which results are main evidence and which are boundary evidence.

\subsection{H100 reasoning fidelity gate}

The most stable core result in the current empirical package is the H100 reasoning fidelity gate. Across all budgets of \textbf{256}, \textbf{384}, and \textbf{512} in both the \textbf{AIME24} and \textbf{AIME25} 6-example slices, CASK records higher top-1 agreement and top-5 coverage than TriAttention. Mean NLL is lower across all ranges in \textbf{AIME24}, and in \textbf{AIME25} shows near-parity at \textbf{256} and advantage at \textbf{384} and \textbf{512}. This result shows that the same-budget advantage recurs across two reasoning slices rather than being a single-witness artifact.

More importantly, the crossing pattern. In \textbf{AIME24}, both \textbf{cask@256 > triattention@384} and \textbf{cask@384 > triattention@512} hold. In \textbf{AIME25}, crossing is weaker but \textbf{cask@384 > triattention@512} is maintained. This result makes clear that CASK's advantage does not lie in discarding more aggressively at the same nominal budget. The actual interpretation is closer to the opposite: even when the saved ratio is similar or CASK is less aggressive, it forms a stronger budget-fidelity frontier capable of competing with higher-budget eviction baselines.

\begin{figure}[H]
\centering
\includegraphics[width=\linewidth]{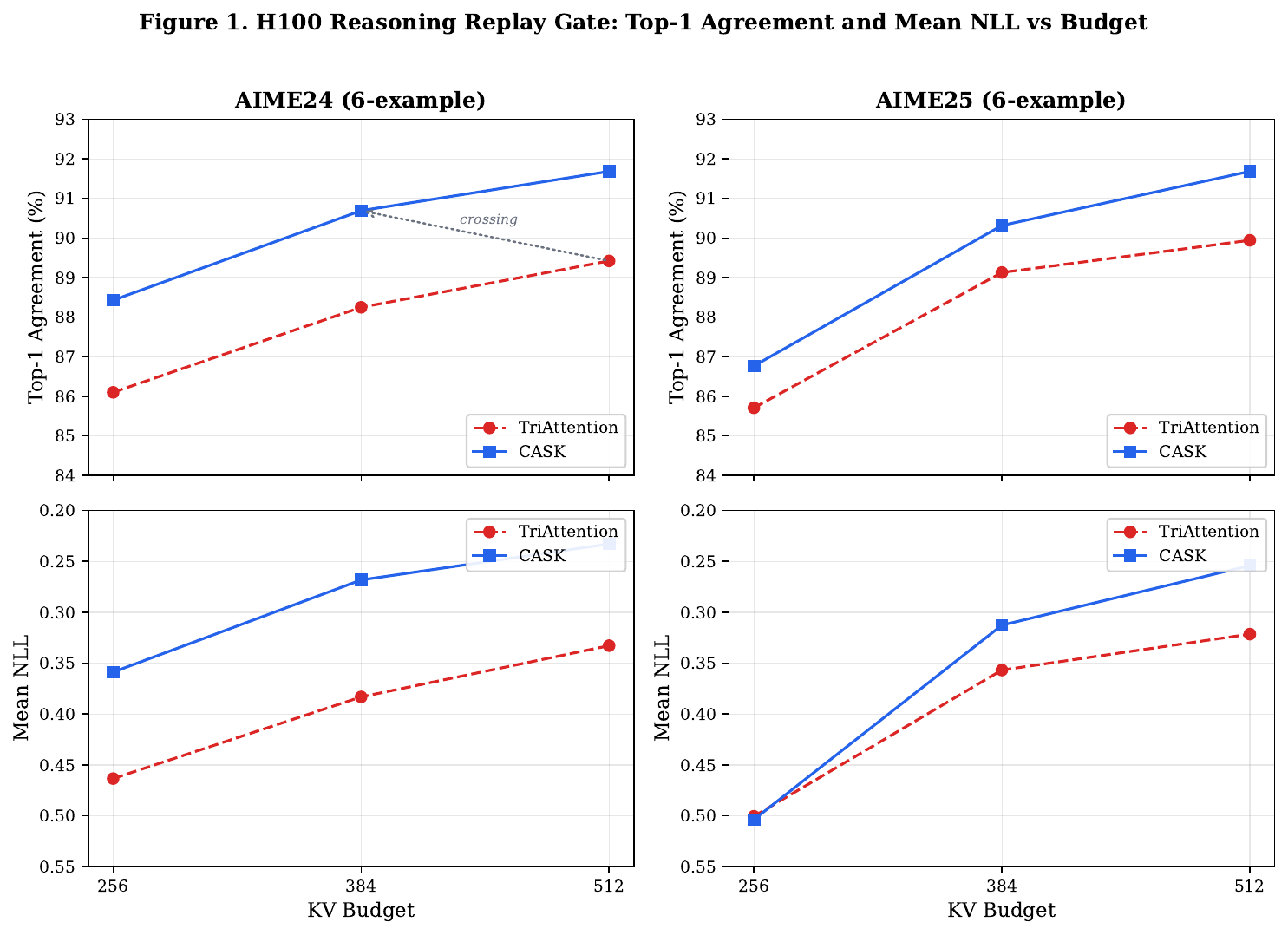}
\caption{H100 reasoning replay gate. CASK improves replay fidelity at matched budgets on both AIME24 and AIME25, and repeatedly crosses higher-budget TriAttention points.}
\label{fig:reasoning-gate}
\end{figure}

The H100 reasoning gate should therefore be read as this paper's mainline evidence. The safest claim from this section is that "CASK repeatedly shows same-budget fidelity advantage and partial budget crossing in reasoning slices." There is no need to claim universal superiority, and current evidence does not demand it. \textbf{Quantitative raw replay tables and measured-count-based audits are in Appendix C.}

\begin{table}[H]
\centering
\small
\caption{Table 1. H100 reasoning fidelity gate}
\begin{tabularx}{\textwidth}{YYYYYYYY}
\toprule
Slice & Method & Budget & Top-1 & Top-5 & Mean NLL & First Mismatch & Saved Ratio \\
\midrule
AIME24 ref6 & TriAttention & 256 & 86.1\% & 98.0\% & 0.463 & 8.7 & 65.3\% \\
AIME24 ref6 & \textbf{CASK} & 256 & \textbf{88.4\%} & \textbf{99.2\%} & \textbf{0.359} & 8.7 & 65.3\% \\
AIME24 ref6 & TriAttention & 384 & 88.2\% & 98.9\% & 0.383 & 8.7 & 61.6\% \\
AIME24 ref6 & \textbf{CASK} & 384 & \textbf{90.7\%} & \textbf{99.7\%} & \textbf{0.268} & 8.7 & 61.6\% \\
AIME24 ref6 & TriAttention & 512 & 89.4\% & 99.1\% & 0.333 & 8.7 & 43.6\% \\
AIME24 ref6 & \textbf{CASK} & 512 & \textbf{91.7\%} & \textbf{99.9\%} & \textbf{0.233} & 8.7 & 43.5\% \\
AIME25 ref6 & TriAttention & 256 & 85.7\% & 97.8\% & \textbf{0.500} & 17.5 & 63.4\% \\
AIME25 ref6 & \textbf{CASK} & 256 & \textbf{86.8\%} & \textbf{97.9\%} & 0.504 & 17.5 & 55.9\% \\
AIME25 ref6 & TriAttention & 384 & 89.1\% & 98.8\% & 0.357 & 18.2 & 59.5\% \\
AIME25 ref6 & \textbf{CASK} & 384 & \textbf{90.3\%} & \textbf{99.1\%} & \textbf{0.313} & 18.2 & 63.3\% \\
AIME25 ref6 & TriAttention & 512 & 89.9\% & 99.0\% & 0.321 & 17.5 & 44.8\% \\
AIME25 ref6 & \textbf{CASK} & 512 & \textbf{91.7\%} & \textbf{99.6\%} & \textbf{0.254} & 17.5 & 37.2\% \\
\bottomrule
\end{tabularx}
\end{table}

\subsection{Prompt-heavy replay: gains, roles, and boundaries}

The prompt-heavy package has more complex results, so it should be interpreted by dividing roles rather than reading from a single wide table. On a length-weighted aggregate basis, CASK shows higher weighted top-1, weighted top-5, and lower weighted mean NLL than TriAttention at both \textbf{256} and \textbf{384} budgets. That is, at the package-wide level, the same-budget replay gain is clear. However, each witness means something different.

\begin{figure}[t]
\centering
\includegraphics[width=\linewidth]{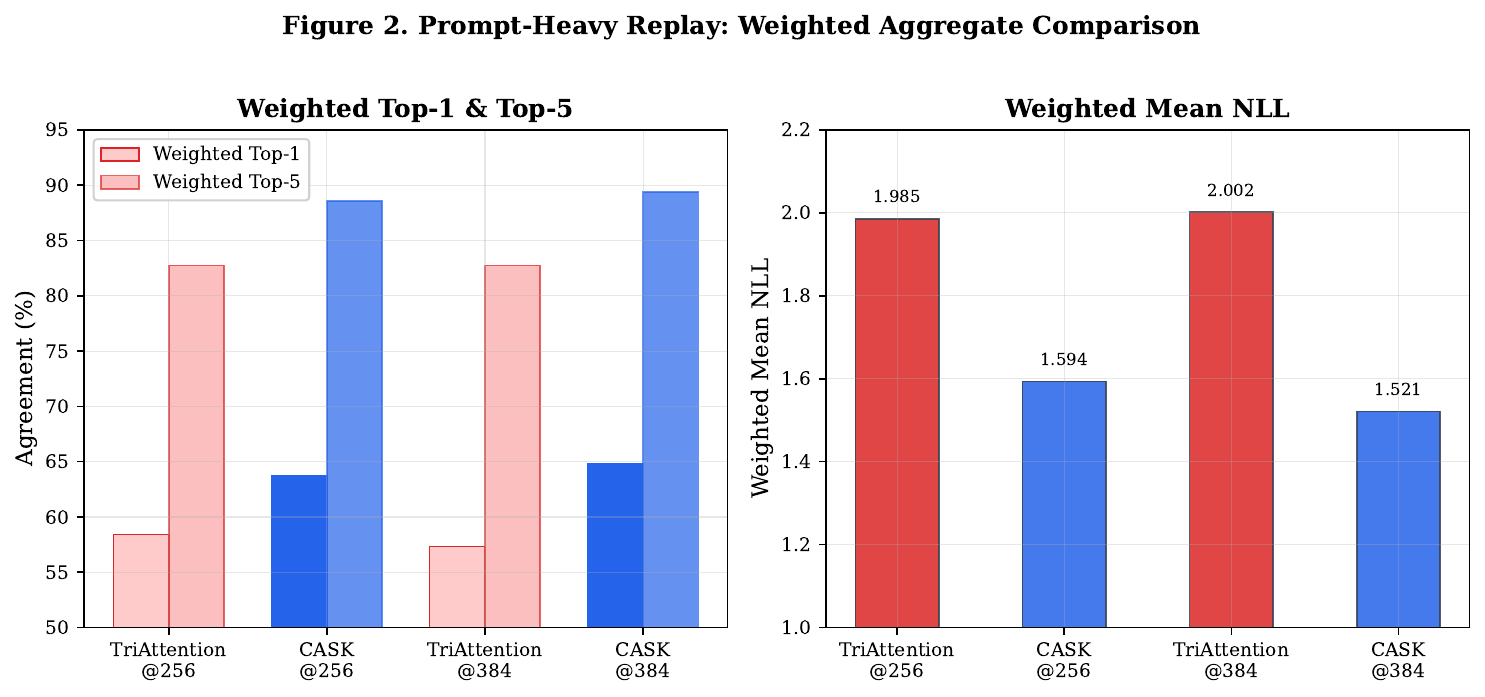}
\caption{Prompt-heavy weighted aggregate. Across the package-level replay summary, CASK improves weighted top-1, weighted top-5, and weighted mean NLL at both budgets.}
\label{fig:promptheavy-aggregate}
\end{figure}

The strongest same-budget witness is \textbf{hotpotqa}. In the current matrix, at \textbf{budget = 256}, CASK recorded top-1 93.8\% and mean NLL 0.151 versus TriAttention's top-1 81.3\% and mean NLL 1.374, and at \textbf{budget = 384}, CASK recorded top-1 96.9\% and mean NLL 0.110 versus TriAttention's top-1 81.3\% and mean NLL 1.344. This witness has very large same-budget fidelity gain and relatively straightforward interpretation.

\textbf{multi\_news} is important in a different sense. This witness is a clean decode-active replay case where decode events are actually recorded. At both same-budget \textbf{256} and \textbf{384}, CASK shows higher top-1, top-5, and lower mean NLL than TriAttention. This means that \textbf{multi\_news} shows the two-stage policy does not merely adjust prefix-side coverage but can also lead to replay fidelity gains in regimes where decode-stage activity actually exists.

Conversely, \textbf{qasper} should not be read the same way. In the current package, \textbf{qasper} has large improvement and is strong as crossing evidence, but should be interpreted as prefix-dominant regime in the replay records. That is, \textbf{qasper} is stage-1 coverage and crossing evidence showing that two-stage design forms a stronger budget-fidelity frontier in prompt-heavy regimes, rather than decode-stage merge superiority evidence. Blurring this distinction exaggerates the method claim.

\textbf{musique} has improvement but is not a headline witness. Same-budget advantage is visible but the effect size is smaller, and it is not clean enough to drive the overall package narrative. \textbf{musique} is therefore best positioned as a "case where gain is weaker but direction is maintained."

Finally, \textbf{2wikimqa} is a retained boundary case. At \textbf{budget = 256}, CASK leads in top-1, top-5, and mean NLL, but at \textbf{budget = 384}, top-1 is not fully flipped. That is, this witness shows that CASK does not achieve clean wins across all prompt-heavy multi-hop QA. However, since top-5 and mean NLL are still improved, it is more accurately read as a coverage-sensitive boundary rather than simply called a failure example.

\begin{table}[H]
\centering
\small
\caption{Table 2. Prompt-heavy weighted aggregate}
\begin{tabularx}{\textwidth}{YYYYY}
\toprule
Method & Budget & Weighted Top-1 & Weighted Top-5 & Weighted Mean NLL \\
\midrule
TriAttention & 256 & 58.42\% & 82.74\% & 1.985 \\
TriAttention & 384 & 57.34\% & 82.74\% & 2.002 \\
\textbf{CASK} & 256 & \textbf{63.72\%} & \textbf{88.59\%} & \textbf{1.594} \\
\textbf{CASK} & 384 & \textbf{64.81\%} & \textbf{89.40\%} & \textbf{1.521} \\
\bottomrule
\end{tabularx}
\end{table}

\begin{table}[H]
\centering
\small
\caption{Table 3. Prompt-heavy same-budget summary}
\begin{tabularx}{\textwidth}{YYYYYYYYY}
\toprule
Task & Budget & Tri Top-1 & CASK Top-1 & Delta Top-1 & Tri Mean NLL & CASK Mean NLL & Delta NLL & Decode Events \\
\midrule
qasper & 256 & 67.2\% & \textbf{71.1\%} & \textbf{+3.9\%p} & 1.315 & \textbf{1.247} & \textbf{-0.068} & 0 \\
qasper & 384 & 66.4\% & \textbf{73.4\%} & \textbf{+7.0\%p} & 1.398 & \textbf{1.297} & \textbf{-0.101} & 0 \\
multi\_news & 256 & 54.7\% & \textbf{60.0\%} & \textbf{+5.3\%p} & 2.060 & \textbf{1.652} & \textbf{-0.408} & 3 \\
multi\_news & 384 & 53.7\% & \textbf{61.3\%} & \textbf{+7.6\%p} & 2.052 & \textbf{1.540} & \textbf{-0.512} & 3 \\
hotpotqa & 256 & 81.3\% & \textbf{93.8\%} & \textbf{+12.5\%p} & 1.374 & \textbf{0.151} & \textbf{-1.223} & 0 \\
hotpotqa & 384 & 81.3\% & \textbf{96.9\%} & \textbf{+15.6\%p} & 1.344 & \textbf{0.110} & \textbf{-1.234} & 0 \\
musique & 256 & 59.4\% & \textbf{65.6\%} & \textbf{+6.2\%p} & \textbf{2.697} & 2.713 & +0.016 & 0 \\
musique & 384 & 53.1\% & \textbf{62.5\%} & \textbf{+9.4\%p} & 2.862 & \textbf{2.650} & \textbf{-0.212} & 0 \\
2wikimqa & 256 & 59.4\% & \textbf{62.5\%} & \textbf{+3.1\%p} & 3.368 & \textbf{2.375} & \textbf{-0.993} & 0 \\
2wikimqa & 384 & \textbf{59.4\%} & 56.3\% & -3.1\%p & 3.415 & \textbf{2.397} & \textbf{-1.018} & 0 \\
\bottomrule
\end{tabularx}
\end{table}

In summary, the honest headline for the prompt-heavy replay package is not "decode-stage merge is strongly activated across all tasks." A more accurate statement is that strong same-budget gains clearly exist, replay-level decode-active witnesses are limited but real, and boundary cases also remain. This bounded profile is precisely what increases the credibility of the current empirical package, because it clearly separates the intervals where evidence is closed and where boundaries still exist. \textbf{Prompt-heavy raw audit, measured counts, and scaling tables are in Appendix C; witness taxonomy and package-level reading guide are in Appendix D.}

\begin{figure}[t]
\centering
\includegraphics[width=\linewidth]{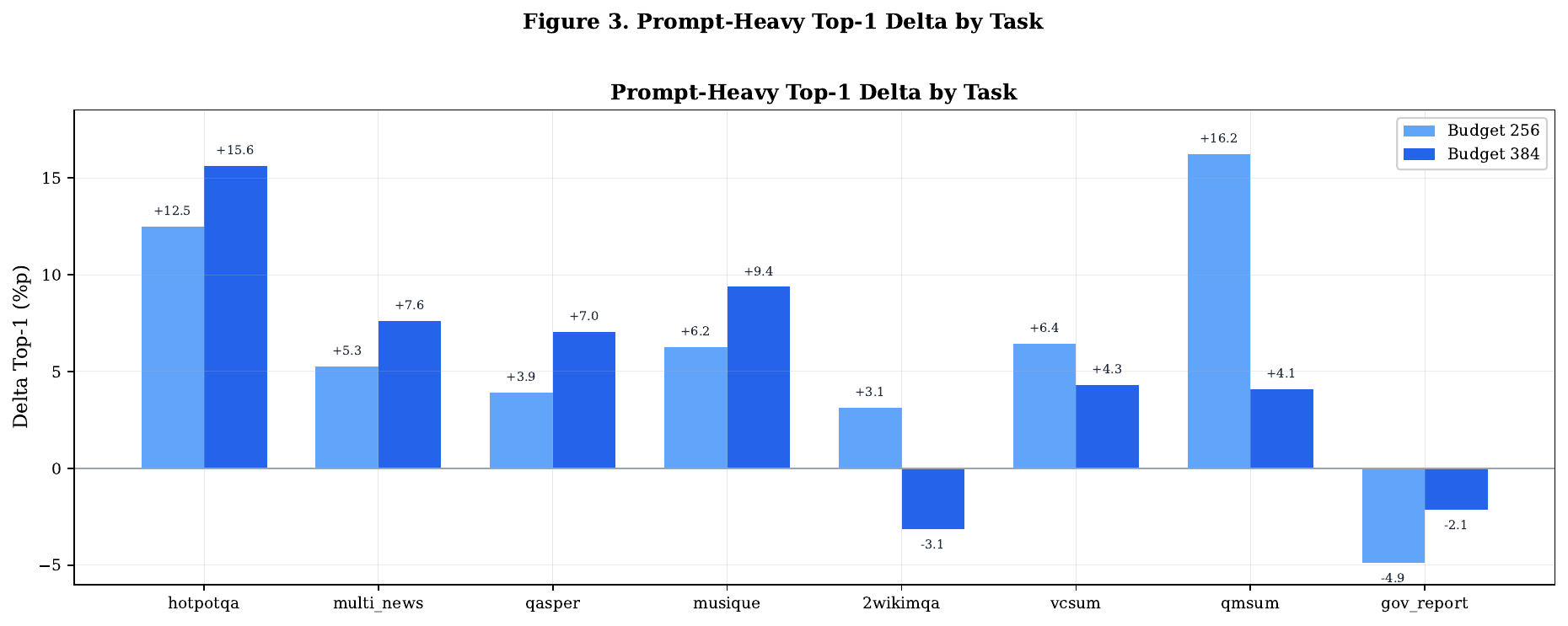}
\caption{Prompt-heavy witness map. The tasks do not all play the same role: some are decode-active witnesses, some are prefix-dominant crossings, and some are retained boundaries.}
\label{fig:witness-map}
\end{figure}

\subsection{Actual-output bridge}

To show that teacher-forced replay is not an entirely separate diagnostic value from actual generation, we maintain the actual-output bridge package separately. This package serves three purposes. First, in \textbf{qasper}, it shows output-level crossing. \textbf{cask@256} records higher sequence ratio and higher task metric than \textbf{triattention@512}. This provides a minimal bridge showing that replay-level crossing is not an artifact entirely unrelated to actual generation.

Second, in \textbf{multi\_news}, same-budget output bridge is confirmed. Under same-budget \textbf{384} conditions, TriAttention effectively collapses, while CASK maintains partial sequence preservation and non-zero task metric. This result means \textbf{multi\_news} is not only a replay-level decode-active witness but also a case showing user-visible difference at the output level.

Third, \textbf{hotpotqa} should be read as a parity non-regression case, not a gain case. In current evidence, CASK maintains identical output quality while performing compression. This too is important. The purpose of the actual-output package is not to claim that CASK produces large gains across all tasks but to honestly present that both gain cases and parity cases exist.

\begin{figure}[t]
\centering
\includegraphics[width=\linewidth]{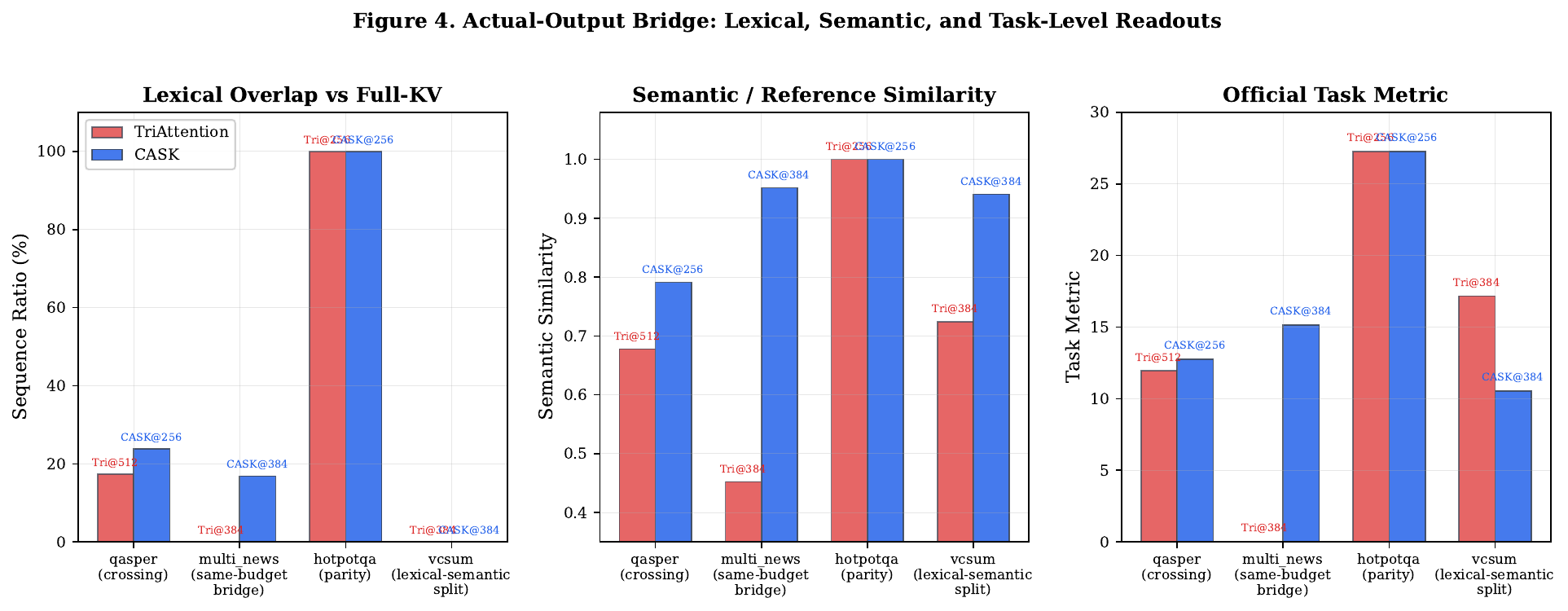}
\caption{Actual-output bridge. Replay-level gains are not purely diagnostic: they connect to lexical, semantic, and task-level output behavior on selected witnesses.}
\label{fig:actual-bridge}
\end{figure}

\begin{table}[H]
\centering
\small
\begin{tabularx}{\textwidth}{YYYYYYYYY}
\toprule
Task & Variant & Budget & Sequence Ratio & Prefix Ratio & Output Ratio & Task Metric & Terminal Saved & Compression Events \\
\midrule
qasper & TriAttention & 512 & 17.3\% & 3.7\% & 100.0\% & 11.94 & - & - \\
qasper & \textbf{CASK} & 256 & \textbf{23.8\%} & \textbf{5.5\%} & 100.0\% & \textbf{12.77} & 90.9\% & 1.0 \\
multi\_news & TriAttention & 384 & 0.0\% & 0.0\% & 100.0\% & 0.00 & - & - \\
multi\_news & \textbf{CASK} & 384 & \textbf{16.9\%} & \textbf{8.1\%} & 100.0\% & \textbf{15.16} & 84.1\% & 3.0 \\
hotpotqa & TriAttention & 256 & 100.0\% & 100.0\% & 100.0\% & 27.27 & - & - \\
hotpotqa & CASK & 256 & 100.0\% & 100.0\% & 100.0\% & 27.27 & 97.6\% & 0.0 \\
\bottomrule
\end{tabularx}
\end{table}

The actual-output bridge should therefore be positioned not as an axis replacing the main results but as an auxiliary axis showing that replay fidelity gains are not entirely separated from actual generation. Maintaining this framing ensures that the entire evidence package interlocks without exaggeration. \textbf{How to read the bridge package within the overall evidence bundle is in Appendix D.}

\subsection{What Section 4 supports}

What Section 4 currently directly supports is threefold. First, in reasoning slices, same-budget fidelity advantage and some budget crossings recur. Second, the prompt-heavy replay package shows strong same-budget gains, but within it, decode-active witnesses, prefix-dominant witnesses, and retained boundaries must be read separately. Third, the actual-output bridge shows that replay-level gains are not entirely unrelated to actual generation.

Conversely, what Section 4 does not yet support is also clear. Universal superiority across all benchmarks, decode-stage dominance across all witnesses, and higher savings across all budgets cannot be claimed with current evidence. The honest headline for the results section should therefore be limited to something like "CASK shows a stronger budget-fidelity frontier within the current package."

\section{Analysis and Discussion}

The most important point when interpreting current results is that CASK's gains should not be read as simple compression aggressiveness. Looking at the reasoning fidelity gate and prompt-heavy replay package together, CASK maintains higher replay fidelity despite being less aggressive or having similar saved ratio to TriAttention in some settings. This again shows that CASK's advantage lies not in "discarding more at the same budget" but in \textbf{lowering the usable budget itself required to maintain the same full-KV behavior}.

This interpretation also connects with Phase 1's conclusions. Phase 1 started from the expectation that designing a more sophisticated scorer would change the core of compression outcomes, but in practice, keep-set churn was limited. By contrast, current results show that the fidelity frontier can move when preserve/consolidate separation is introduced at the policy level. That is, the current package more strongly supports the hypothesis that "role differentiation in reasoning traces is the key lever" over the hypothesis that "a better score family is everything." \textbf{Detailed narrative of Phase 1 and formula archive are in Appendix A and Appendix B respectively.}

\subsection{Why protected core alone is not enough}

An important point revealed in internal diagnostics is that protected core alone is not sufficient. In representative healthy witnesses, protected core alone does not directly preserve all oracle top-k mass. Yet the final representative set recovers a substantial portion of it. This observation shows that CASK's purpose is not to keep only the core and discard the rest. In practice, core preservation and representative reconstruction must work together to maintain high fidelity.

\begin{figure}[t]
\centering
\includegraphics[width=\linewidth]{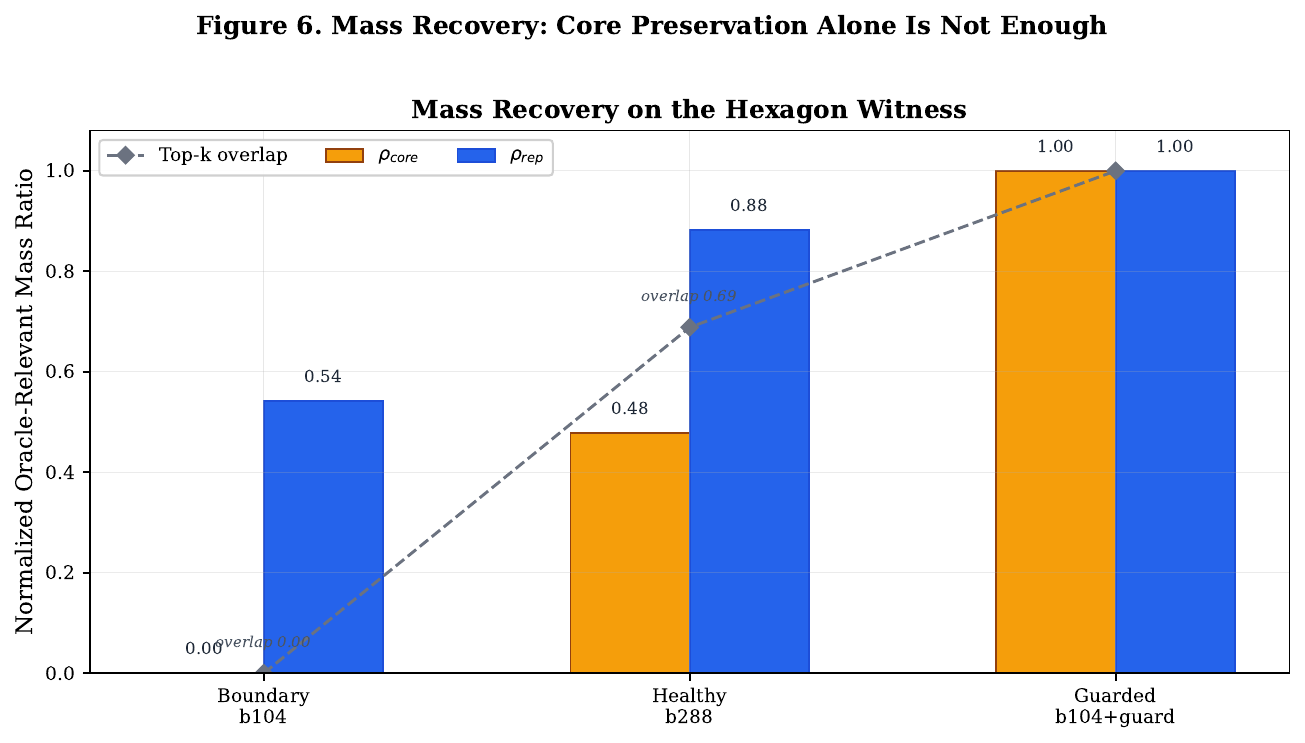}
\caption{Mass recovery on the hexagon witness. Directly protected core mass is insufficient on its own; representative reconstruction recovers a large share of oracle-relevant mass.}
\label{fig:mass-recovery}
\end{figure}

This point makes the method claim more precise. CASK's advantage does not lie in the simple proposition "core is important." A more accurate statement is that within reasoning traces, there exists a core that must be excluded from merge, and the scratch outside can recover reference-relevant mass through the representative set. That is, a healthy regime is created only when preserve and reconstruct work together.

\subsection{Healthy regime versus collapse regime}

Current evidence also suggests that quality cliff is difficult to explain with a single similarity drop. In healthy regimes, representative mass ratio is high, local merge occurs within relatively stable temporal neighborhoods, and top-k overlap and NLL degradation are maintained gradually. By contrast, in collapse regimes, when the prefix encroaches on the budget, available slots are excessively reduced, or the representative set fails to recover sufficient reference-relevant mass, sharp fidelity loss appears.

\begin{figure}[t]
\centering
\includegraphics[width=\linewidth]{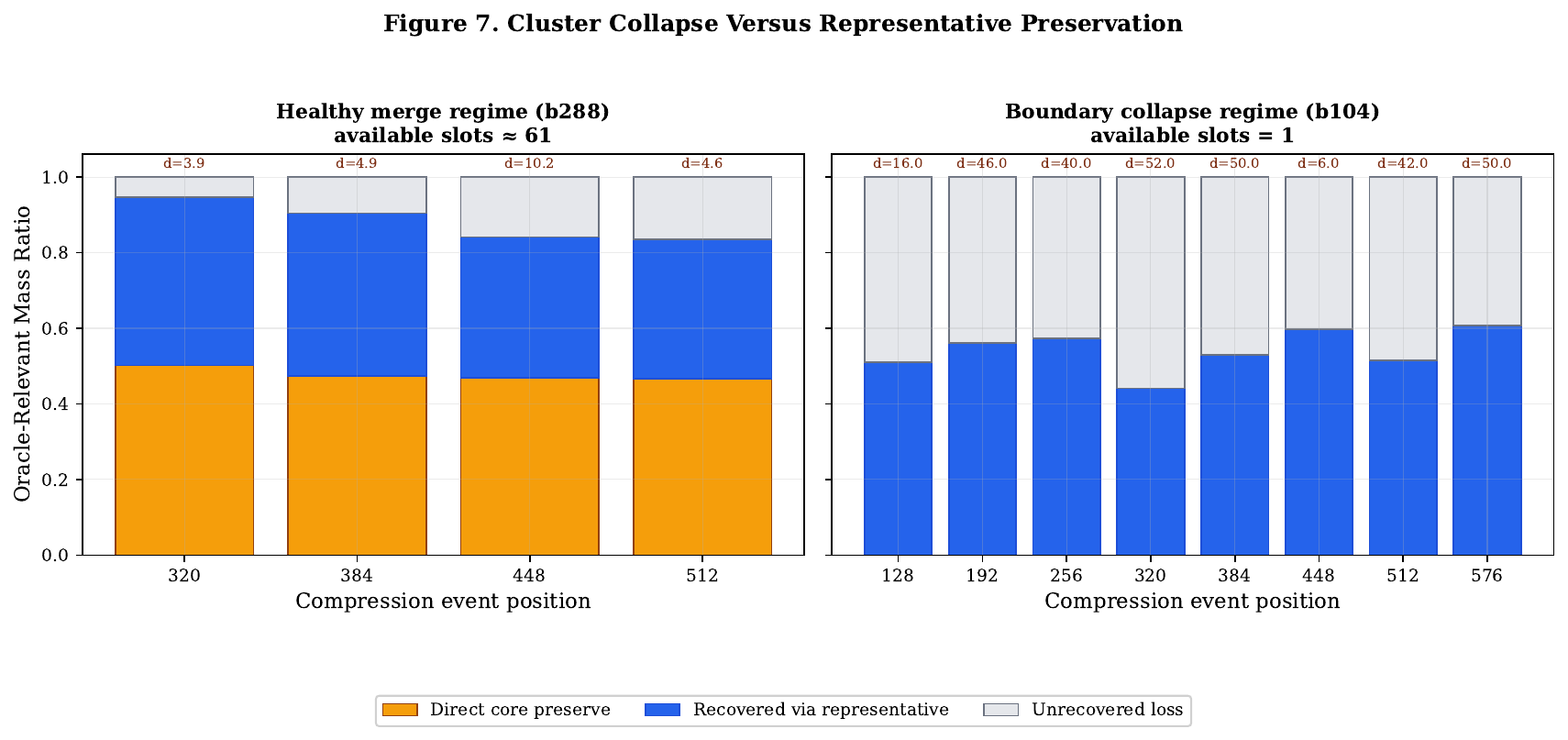}
\caption{Healthy representative preservation versus boundary collapse. Collapse is structural: when available slots collapse, unrecovered oracle-relevant mass grows sharply.}
\label{fig:cluster-collapse}
\end{figure}

This distinction also explains why regime guard is necessary. If \textbf{prefix\_budget\_exhausted} or effectively \textbf{merge\_inactive} states are mixed with active compression evidence, it becomes unclear where the method actually works and where it does not. The reason for treating \textbf{multi\_news} and \textbf{vcsum} separately as decode-active witnesses and separating \textbf{qmsum} and \textbf{gov\_report} as boundary probes in the current prompt-heavy package lies here.

\subsection{Interpreting prompt-heavy gains without overclaiming}

Prompt-heavy results are particularly easy to overstate. Even if both \textbf{qasper} and \textbf{multi\_news} look good, the two witnesses are not the same kind of evidence. \textbf{multi\_news} is a replay-level witness with decode-stage activity actually recorded, and it also has an output-level bridge. By contrast, \textbf{qasper} is strong as crossing evidence but belongs to a prefix-dominant regime. Ignoring this difference and grouping both as "decode-stage merge superiority" exaggerates the claim.

The most honest current interpretation is as follows. The two-stage policy as a whole is clearly effective in prompt-heavy regimes. However, clean witnesses where decode-stage consolidation is front-and-center are limited, with \textbf{multi\_news} and \textbf{vcsum} serving as the current replay-level clean witnesses and \textbf{multi\_news} serving as the core output-level bridge witness. This profile is not a weakness but rather a strength of claim discipline, because it clearly delineates the intervals where evidence is closed and where boundaries remain.

\subsection{Why the headline is minimum usable budget}

Reading current results as a fixed-budget savings competition blurs the paper's center. CASK may not show higher savings than TriAttention in some slices, or may even be less aggressive. Nevertheless, if same-budget fidelity is higher and crossing occurs with higher-budget baselines, the center of interpretation should be the usable budget frontier, not the savings ratio.

This framing is particularly important for reasoning compression. In reasoning workloads, how long answer-forming behavior can be maintained within a given budget matters more than simply reducing cache token counts. CASK's strongest message is not that it is the most aggressive at a nominal budget but that it \textbf{lowers the behavior-preserving minimum budget further down}.

\subsection{What remains unresolved}

Simultaneously, unresolved points are clear. First, the current package is beyond single-witness level but still not a broad matrix covering an entire full benchmark sweep. Second, output-level bridge is limited to some witnesses, with clean decode-active output bridge effectively having \textbf{multi\_news} as core. Third, coverage-sensitive multi-hop QA like \textbf{2wikimqa} shows that CASK is not a policy that always achieves clean wins.

These unresolved points are not elements that weaken the current paper but rather boundaries that indicate how far to claim and where to leave as future work. That is, the current package is sufficiently strong, but its strength lies in well-bounded, well-interpreted evidence, not universal dominance.

\section{Related Work}

CASK's positioning is clear along two axes. First, the relationship with eviction-centered reasoning KV compression. Lines such as TriAttention, R-KV, and Crystal-KV all center the compression primitive on discard, providing important design elements including reasoning-friendly scoring, segment-aware weighting, and answer-think asymmetry, but still treat compression as the problem of whether to keep or discard individual tokens \citep{mao2026triattention,cai2025rkv,wang2026crystalkv}. CASK shares this perspective but is distinguished by treating preserve and consolidate as different primitives rather than viewing the problem as a ranking-only problem.

Second, the relationship with merge-oriented compression. KeepKV, KVMerger, AsymKV, CaM, and D2O provide important foundations from the consolidation operator perspective \citep{tian2025keepkv,wang2024kvmerger,qian2025asymkv,zhang2024cam,wan2024d2o}. CASK is closer to this line but is distinguished by centering on \textbf{what to protect and what to consider as merge targets in reasoning traces} rather than the general theory of the merge operator itself. That is, CASK can be positioned as a method sitting atop the generic merge family while foregrounding reasoning-specific policy.

Third, from the perspective of two-stage compression design, CASK occupies a unique position. Unlike existing approaches that treat prefix and decode with the same primitive, CASK introduces structural separation applying eviction to the prefix and selective consolidation to the decode. This design is an essential component for securing method coverage in prompt-heavy regimes and creates the conditions for the decode-stage policy to actually activate.

\section{Limitations and Claim Boundary}

The current empirical package has already moved beyond the single-witness level, but this does not mean it has closed an entire broad benchmark sweep. This paper should therefore state clearly what is directly supported at present and what remains at the boundary, rather than overstating the results. The purpose of this section is not to hide weaknesses but to fix claim discipline and increase the credibility of result interpretation.

\subsection{What the current package directly supports}

The claims directly supported by current evidence should be most honestly limited to the following. First, in reasoning slices, CASK repeatedly shows same-budget fidelity advantage. In the \textbf{AIME24} and \textbf{AIME25} reasoning gates, superiority in top-1 agreement, top-5 coverage, and mean NLL recurs, with budget crossing also observed in some ranges. Second, strong same-budget gains exist in the prompt-heavy replay package. However, these gains do not manifest the same way across tasks, and decode-active witnesses and prefix-dominant witnesses must be read separately. Third, the actual-output bridge shows that replay-level fidelity gains are not an artifact entirely separated from actual generation.

That is, the safest claim this paper can make is that CASK shows a stronger budget-fidelity frontier within the current package. This claim is sufficiently strong while remaining bounded. Universal dominance or across-the-board superiority is therefore not claimed.

\subsection{What remains outside the current claim}

Conversely, claims not yet directly supported by current evidence are also clear. First, consistent absolute accuracy superiority across all reasoning benchmarks is not yet guaranteed. Second, one cannot claim decode-stage merge is front-and-center across all witnesses. Third, one cannot say higher fixed-budget savings are shown across all budgets. Fourth, one cannot conclude that the current correction set or reserve schedule is already optimal.

\subsection{Prompt-heavy regime must be split by regime}

The prompt-heavy package particularly requires claim discipline. The current replay-level decode-active witnesses should be read as the two of \textbf{multi\_news} and \textbf{vcsum}. In contrast, \textbf{qasper} is very strong as crossing evidence but should be interpreted as prefix-dominant regime. \textbf{qmsum} and \textbf{gov\_report} are probes revealing the \textbf{prefix\_budget\_exhausted} boundary and are not evidence for asserting the method's universal superiority. \textbf{2wikimqa} remains as a retained boundary case.

\subsection{Output-level bridge is still limited}

The actual-output bridge is also currently limited to some witnesses. \textbf{qasper} is an output-level crossing case, \textbf{multi\_news} is a same-budget bridge case, and \textbf{hotpotqa} is a parity non-regression case. However, this does not immediately mean "replay gain always translates to output gain across broad benchmarks." The output-level bridge is currently important auxiliary evidence but not yet a large-scale closed package.

\subsection{Phase 1 and Phase 2 are not competing stories}

Another point to clarify is that Phase 1 and the current CASK narrative are not competing with each other. Phase 1 explored the potential of scorer-side refinement and left behind the horizon kernel language and diagnostic archive. Yet it also revealed the limited nature of actual keep-set churn. Current CASK is a policy-level pivot that departed precisely from that limitation. Phase 1 is therefore not a discarded past but an analytical predecessor that explains why-not-scorer and justifies the current claim boundary. \textbf{More detailed account of this linkage is in Appendix A, Phase 1 formula assets in Appendix B, raw replay audit in Appendix C, and how to read the entire empirical package in Appendix D.}

\subsection{Asset provenance and license handling}

This paper uses external models, benchmarks, and helper code, but it does not relicense those assets under a new umbrella license. The paper-facing repository itself is released under \textbf{Apache-2.0}. The evaluated \textbf{Qwen3-8B} model is obtained from its upstream Hugging Face release, whose model card lists the license as \textbf{Apache-2.0} (\url{https://huggingface.co/Qwen/Qwen3-8B}), and the model weights are \textbf{not} redistributed in this repository. The long-context benchmark suite used for prompt-heavy evaluation is based on \textbf{LongBench}, whose upstream repository is released under the \textbf{MIT License} (\url{https://github.com/THUDM/LongBench}), and benchmark data remain subject to the upstream benchmark terms. The bundled \texttt{latex2sympy} helper in our evaluation tree also retains its own \textbf{MIT License}.

The operational rule is simple: we release only our own method code, scripts, tables, figure packs, and paper sources, while users must obtain external models and datasets from their original sources under the original terms of use. This separation matters for both reproducibility and compliance. It prevents paper artifacts from silently repackaging third-party assets while still making the experimental path fully traceable.

\subsection{Broader impacts}

This work has both positive and negative implications. On the positive side, lowering the usable KV-cache budget can reduce the memory cost of long-context and reasoning inference, which can make evaluation and deployment accessible on smaller hardware and lower the energy and cost required to study long-context behavior at fixed model size.

The same efficiency gain also lowers the cost of deploying capable long-reasoning systems more broadly. In the intended setting, compression errors can silently alter intermediate reasoning traces, so deployment-side savings may come with hidden behavior drift if the method is used outside the validated regime. In the misuse setting, cheaper long-context inference can modestly reduce the barrier to scaling systems that produce misleading, manipulative, or otherwise harmful content. Our mitigation is limited but concrete: we do not release a new base model or a new dataset, we keep claim boundaries explicit, and we expose replay-fidelity audits and failure boundaries instead of presenting the method as universally safe or universally dominant.

\subsection{Final claim discipline}

The claim discipline of this paper should therefore be fixed by the following statement. CASK's strength does not lie in always reducing more, always showing higher accuracy, or always showing stronger decode-stage merge. A more accurate expression is that \textbf{the policy design that distinguishes protected core from mergeable scratch within reasoning traces creates a stronger budget-fidelity frontier within the current evidence package, and consequently can lower the usable budget required to maintain the same full-KV behavior}.

\section{Conclusion}

This paper resets the central question of reasoning KV compression. The problem does not lie solely in more precisely predicting which token is more important. The more fundamental question is which tokens must be preserved and which can be safely consolidated. From this perspective, we proposed CASK, which decomposes the reasoning trace into protected core and mergeable scratch, preserves the core, and performs selective consolidation only on the scratch. Additionally, in prompt-heavy regimes, we restructured the design through a two-stage policy separating prefix eviction and decode consolidation to prevent method coverage itself from collapsing.

The conclusion most strongly supported by current evidence is that CASK can show same-budget fidelity advantage, and its core value lies not in absolute superiority of fixed-budget savings but in lowering the minimum usable budget frontier. In the reasoning gate, same-budget superiority and partial budget crossings recurred, and in the prompt-heavy replay package, strong gains, limited but real decode-active witnesses, and retained boundary cases were jointly observed. The actual-output bridge also shows that replay-level fidelity gains are not an artifact entirely separated from actual generation.

Simultaneously, this paper does not make claims beyond the scope of current evidence. CASK does not aim for universal superiority across all benchmarks or always more aggressive compression. Rather, what the current package directly supports is that reflecting role differentiation of reasoning traces in policy can form a stronger budget-fidelity frontier. This conclusion is also consistent with the process of pushing scorer-side refinement in Phase 1, confirming the limited nature of actual keep-set churn, and pivoting to preserve/consolidate separation.

The most honest conclusion of this paper is therefore as follows. Reasoning KV compression should be treated not as a ranking-only problem but as a behavior-preserving structured consolidation problem. Within the current evidence package, CASK demonstrates that this directional shift can lead to substantive fidelity gains and a lower minimum usable budget.

\appendix

\section{Phase 1 Background and Pivot Rationale}

Phase 1 is not a discarded attempt separate from CASK's main text but an analytical predecessor showing why scorer-side refinement alone was insufficient. This paper therefore does not treat Phase 1 as a simple historical note but preserves it as the logical basis for the current directional pivot.

\subsection{What Phase 1 tried to do}

Phase 1's basic hypothesis was that the key to reasoning KV compression lies in designing better scoring rules. To this end, scorer-side refinements such as adaptive horizon, RMS2, and variational horizon were explored, and score families were interpreted through the horizon kernel language. Two concerns existed: whether scoring rules that better reflect future relevance exist, and whether those scoring rules reorganize the actual keep-set to better preserve reasoning behavior.

\subsection{What Phase 1 was trying to prove}

What Phase 1 substantively wanted to prove was not simply that "the new scorer is mathematically more plausible." The stronger goal was to show that scorer refinement is the primary lever that changes actual compression outcomes. That is, to show that mathematical differences in score families lead to actual differences in token survival patterns and representative quality, and consequently that same-budget fidelity is systematically improved.

\subsection{What Phase 1 actually found}

What emerged from Phase 1's exploration, however, was more complex. The horizon kernel, RMS2 decomposition, and variational horizon formulation were meaningful as diagnostic language and mathematical archive, but actual keep-set churn was more limited than expected. Even when the scorer changed, the set of protected tokens did not substantially reorganize, and differences in kernel-space did not directly translate to behavior-level differences. In other words, the existence of mathematical refinement itself did not sufficiently demonstrate being the key lever of compression policy.

\subsection{Why this forced a pivot}

This result became the core basis for the directional pivot. If scorer refinement cannot substantially change the actual keep-set, then the bottleneck of reasoning KV compression may not lie in creating better scalar rankings. A more valid question then is not "who is more important" but what tokens to consider as preservation targets and what tokens to view as consolidatable targets in the first place. CASK's current protected core / mergeable scratch decomposition departs precisely from this problem redefinition.

\subsection{What remains useful from Phase 1}

That said, Phase 1 was not meaningless. The horizon kernel language, score-family interpretations, frequency-aware diagnostics, and perturbation perspectives obtained in this stage still remain useful as diagnostic language in the current paper. Their role, however, is closer to appendix-level mathematical archive and analysis support than to the method headline. That is, Phase 1 was not the final answer but an essential intermediate step that explains why-not-scorer and justifies CASK's policy-level transition.

\section{Phase 1 Mathematical Archive}

The mathematical assets obtained in Phase 1 do not directly constitute the current main-text headline, but they are needed to explain why-not-scorer and preserve diagnostic language. The formulas below are therefore maintained as an appendix-level archive.

\subsection{Horizon kernel}

$$
\kappa_{\pi_h}(\omega)
:=
\mathbb{E}_{\delta \sim \pi_h}
\left[e^{i\omega\delta}\right]
$$

\subsection{Horizon-averaged mean score}

$$
S^{\mathrm{mean}}_{h,\pi}(k, \Delta)
=
\Re
\sum_f
\mu_{h,f}
\overline{k_{h,f}}
\kappa_{\pi_h}(\omega_f)
e^{i\omega_f \Delta}
$$

\subsection{RMS2 decomposition}

$$
\alpha_f^{\mathrm{RMS2}}
=
\frac{\mathbb{E}\|q_f\|^2 - \|\mu_f\|^2}{\mathbb{E}\|q_f\| + \|\mu_f\|}
=
\alpha_f^{\mathrm{Tri}}
+
\frac{\mathrm{Var}(\|q_f\|)}{\mathbb{E}\|q_f\| + \|\mu_f\|}
$$

\subsection{Variational horizon QP}

$$
\pi_h^{\star}
=
\arg\min_{\pi \in \Delta(G)}
\left\|W_h^{1/2}(B\pi - \tau_h)\right\|_2^2
$$

\subsection{Diagnostic perturbation decomposition}

The representative diagnostic language reads representative perturbation as the sum of within-group dispersion and lost mass.

$$
\left|\sum_{i \in G} a_i \langle q, k_i \rangle - \langle q, \tilde{k}_G \rangle\right|
\lesssim
\|q\|_{\kappa,*} \sum_{i \in G} a_i \|k_i - \tilde{k}_G\|_{\kappa}
+
|\Delta m_G|
$$

The first term reads as within-group $\kappa$-dispersion, and the second as lost mass. In the current paper, this decomposition is used not as a main theorem but as diagnostic language explaining healthy and collapse regimes.

\section{Archival Replay Tables}

This appendix is a space for preserving raw replay audits removed from the main text. The tables below are not tables constituting the main text claim's headline but are needed for appendix-level transparency and audit trail.

Raw experiment data are intentionally preserved outside the main PDF tables. For each headline package, the repository keeps three layers in sync: (1) compact paper-facing summaries under \texttt{artifacts/}, (2) command provenance in \texttt{artifacts/COMMAND\_MAP.md} and package manifests, and (3) raw JSON/CSV/eval/reference roots under \texttt{experiments/frontier/}, \texttt{experiments/outputs/}, and \texttt{experiments/longbench\_h100\_*}. The role of Appendix C is therefore not to duplicate every raw file inside the PDF, but to preserve the measured counts and audit matrices needed to read the claim honestly while keeping the full raw trail accessible and citable from the repository.

\subsection{Raw provenance map}

The main paper-facing packages and their raw back-pointers are:

\begin{itemize}
\item \textbf{Reasoning replay gate}: packaged summaries in \texttt{artifacts/h100\_2026\_04\_10/cask\_h100\_fidelity/}, command manifests in \texttt{experiments/frontier/Qwen3-8B/h100\_aime24\_fidelity\_gate\_20260410/} and \texttt{.../h100\_aime25\_fidelity\_gate\_20260410/}, and raw full-KV references in \texttt{experiments/outputs/aime24/...} and \texttt{experiments/outputs/aime25/...}.
\item \textbf{Prompt-heavy replay package}: packaged readouts in \texttt{artifacts/h100\_2026\_04\_11/}, per-task replay manifests in \texttt{experiments/frontier/Qwen3-8B/h100\_promptheavy\_twostage\_rerun\_20260411\_*}, and raw reference/output trees in \texttt{experiments/longbench\_h100\_refs/}.
\item \textbf{Actual-output bridge}: packaged summaries in \texttt{artifacts/h100\_2026\_04\_11/cask\_h100\_actual\_bridge/}, bridge metrics in \texttt{experiments/frontier/Qwen3-8B/h100\_actual\_bridge\_metrics\_20260411/}, and raw generation/eval trees in \texttt{experiments/longbench\_h100\_actual\_bridge\_20260411/}.
\end{itemize}

In practice, a reader should start from the packaged summary row, follow its \texttt{source\_json} or \texttt{source\_eval\_json} field, and then use \texttt{artifacts/COMMAND\_MAP.md} to recover the exact script, manifest, and raw directory that generated it.

\begin{figure}[t]
\centering
\includegraphics[width=\linewidth]{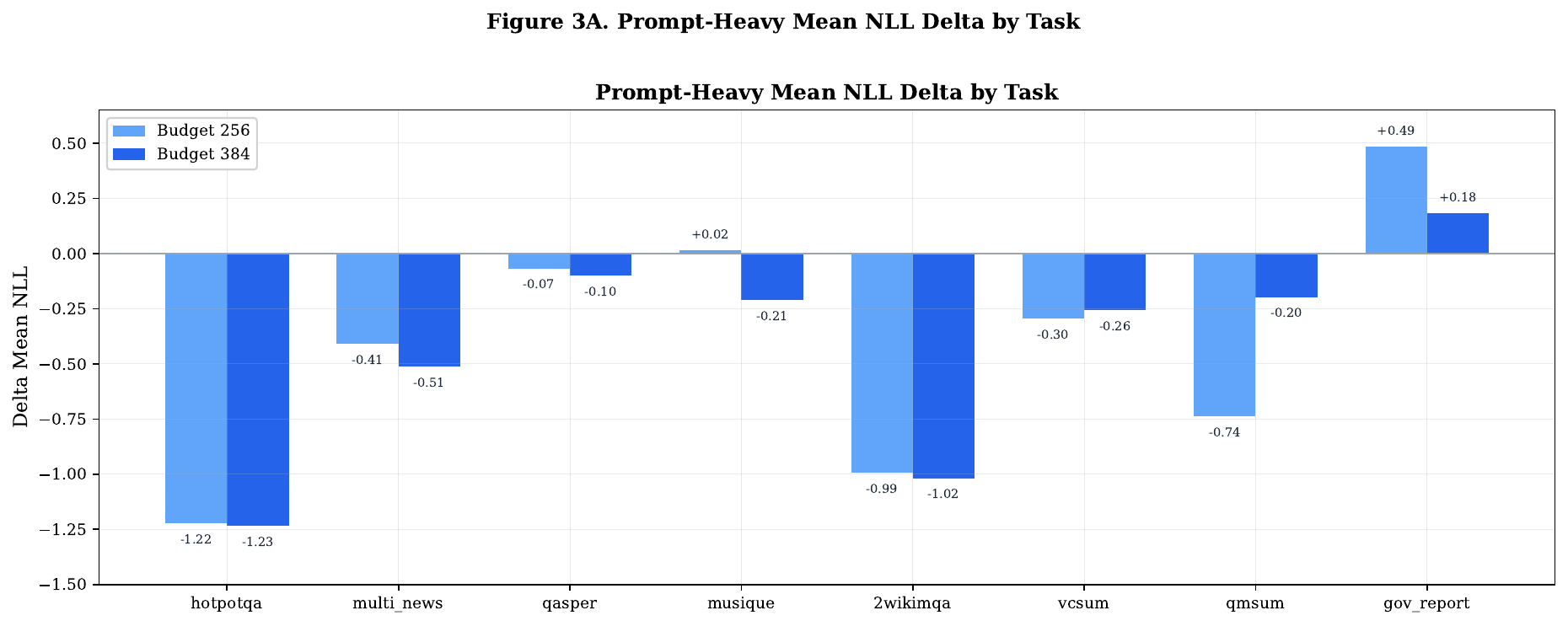}
\caption{Prompt-heavy per-task mean NLL view. This appendix-level figure preserves the per-task NLL readout underlying the package-level prompt-heavy interpretation.}
\end{figure}

\subsection{Weighted aggregate measured counts}

\begin{table}[H]
\centering
\small
\begin{tabularx}{\textwidth}{YYYYYY}
\toprule
Method & Budget & Top-1 Matches & Top-5 Matches & Total Replay Tokens & Weighted Mean NLL \\
\midrule
TriAttention & 256 & 430 & 609 & 736 & 1.985 \\
TriAttention & 384 & 422 & 609 & 736 & 2.002 \\
CASK & 256 & 469 & 652 & 736 & 1.594 \\
CASK & 384 & 477 & 658 & 736 & 1.521 \\
\bottomrule
\end{tabularx}
\end{table}

\subsection{Same-budget measured counts}

\begin{table}[H]
\centering
\small
\begin{tabularx}{\textwidth}{YYYYYYYYY}
\toprule
Dataset & Budget & Output Tokens & Tri Top-1 Matches & CASK Top-1 Matches & Tri Top-5 Matches & CASK Top-5 Matches & Tri First Mismatch & CASK First Mismatch \\
\midrule
qasper & 256 & 128 & 86 & 91 & 115 & 120 & 2 & 4 \\
qasper & 384 & 128 & 85 & 94 & 116 & 118 & 2 & 4 \\
multi\_news & 256 & 512 & 280 & 307 & 420 & 450 & 2 & 2 \\
multi\_news & 384 & 512 & 275 & 314 & 418 & 458 & 2 & 2 \\
hotpotqa & 256 & 32 & 26 & 30 & 29 & 32 & 2 & 11 \\
hotpotqa & 384 & 32 & 26 & 31 & 29 & 32 & 2 & 11 \\
musique & 256 & 32 & 19 & 21 & 23 & 24 & 2 & 3 \\
musique & 384 & 32 & 17 & 20 & 24 & 24 & 2 & 3 \\
2wikimqa & 256 & 32 & 19 & 20 & 22 & 26 & 2 & 2 \\
2wikimqa & 384 & 32 & 19 & 18 & 22 & 26 & 2 & 2 \\
\bottomrule
\end{tabularx}
\end{table}

\subsection{Budget scaling measured counts}

\begin{table}[H]
\centering
\small
\begin{tabularx}{\textwidth}{YYYYYYY}
\toprule
Dataset & Method & Output Tokens & Top-1 Match Count Delta & Top-5 Match Count Delta & Mean NLL Delta & First Mismatch \\
\midrule
qasper & TriAttention & 128 & -1 & +1 & +0.083 & 2 -> 2 \\
qasper & CASK & 128 & +3 & -2 & +0.050 & 4 -> 4 \\
multi\_news & TriAttention & 512 & -5 & -2 & -0.008 & 2 -> 2 \\
multi\_news & CASK & 512 & +7 & +8 & -0.112 & 2 -> 2 \\
hotpotqa & TriAttention & 32 & 0 & 0 & -0.031 & 2 -> 2 \\
hotpotqa & CASK & 32 & +1 & 0 & -0.041 & 11 -> 11 \\
musique & TriAttention & 32 & -2 & +1 & +0.165 & 2 -> 2 \\
musique & CASK & 32 & -1 & 0 & -0.063 & 3 -> 3 \\
2wikimqa & TriAttention & 32 & 0 & 0 & +0.047 & 2 -> 2 \\
2wikimqa & CASK & 32 & -2 & 0 & +0.022 & 2 -> 2 \\
\bottomrule
\end{tabularx}
\end{table}

\subsection{Raw audit matrix}

\begin{table}[H]
\centering
\small
\begin{tabularx}{\textwidth}{YYYYYYYYY}
\toprule
Task & Method & Budget & Top-1 & Top-5 & Mean NLL & First Mismatch & Saved Ratio & CASK Compression Events \\
\midrule
qasper & TriAttention & 256 & 67.2\% & 89.8\% & 1.315 & 2 & 90.9\% & - \\
qasper & TriAttention & 384 & 66.4\% & 90.6\% & 1.398 & 2 & 87.9\% & - \\
qasper & \textbf{CASK} & 256 & \textbf{71.1\%} & \textbf{93.8\%} & \textbf{1.247} & \textbf{4} & 90.9\% & 0 \\
qasper & \textbf{CASK} & 384 & \textbf{73.4\%} & \textbf{92.2\%} & \textbf{1.297} & \textbf{4} & 87.9\% & 0 \\
multi\_news & TriAttention & 256 & 54.7\% & 82.0\% & 2.060 & 2 & 88.1\% & - \\
multi\_news & TriAttention & 384 & 53.7\% & 81.6\% & 2.052 & 2 & 84.1\% & - \\
multi\_news & \textbf{CASK} & 256 & \textbf{60.0\%} & \textbf{87.9\%} & \textbf{1.652} & 2 & 88.1\% & 3 \\
multi\_news & \textbf{CASK} & 384 & \textbf{61.3\%} & \textbf{89.5\%} & \textbf{1.540} & 2 & 84.1\% & 3 \\
hotpotqa & TriAttention & 256 & 81.3\% & 90.6\% & 1.374 & 2 & 97.6\% & - \\
hotpotqa & TriAttention & 384 & 81.3\% & 90.6\% & 1.344 & 2 & 96.5\% & - \\
hotpotqa & \textbf{CASK} & 256 & \textbf{93.8\%} & \textbf{100.0\%} & \textbf{0.151} & \textbf{11} & 97.6\% & 0 \\
hotpotqa & \textbf{CASK} & 384 & \textbf{96.9\%} & \textbf{100.0\%} & \textbf{0.110} & \textbf{11} & 96.5\% & 0 \\
musique & TriAttention & 256 & 59.4\% & 71.9\% & 2.697 & 2 & 98.3\% & - \\
musique & TriAttention & 384 & 53.1\% & 75.0\% & 2.862 & 2 & 97.5\% & - \\
musique & \textbf{CASK} & 256 & \textbf{65.6\%} & 75.0\% & 2.713 & \textbf{3} & 98.3\% & 0 \\
musique & \textbf{CASK} & 384 & \textbf{62.5\%} & 75.0\% & \textbf{2.650} & \textbf{3} & 97.5\% & 0 \\
2wikimqa & TriAttention & 256 & 59.4\% & 68.8\% & 3.368 & 2 & 96.1\% & - \\
2wikimqa & TriAttention & 384 & 59.4\% & 68.8\% & 3.415 & 2 & 94.4\% & - \\
2wikimqa & \textbf{CASK} & 256 & \textbf{62.5\%} & \textbf{81.3\%} & \textbf{2.375} & 2 & 96.1\% & 0 \\
2wikimqa & CASK & 384 & 56.3\% & \textbf{81.3\%} & \textbf{2.397} & 2 & 94.4\% & 0 \\
\bottomrule
\end{tabularx}
\end{table}

\section{Evaluation Map and Reading Guide}

The purpose of this appendix is to organize the current empirical evaluation so that it can be read as a single coherent evidence structure. The current results are not well summarized by a single benchmark average; rather, the reasoning gate, prompt-heavy replay, actual-output bridge, and boundary probes each serve different roles. Appendix D therefore specifies what each result cluster is evidence for, what claims it directly supports, and where boundaries must remain.

\subsection{Main text evidence map}

The tables and figures directly constructing headlines in the current main text are organized into four clusters.

\begin{enumerate}
\item \textbf{H100 reasoning fidelity gate}: Core evidence that same-budget fidelity advantage and partial budget crossing recur in reasoning slices.
\item \textbf{Prompt-heavy weighted aggregate}: Package-level evidence that CASK shows higher weighted top-1, weighted top-5, and lower weighted mean NLL at the overall prompt-heavy package level.
\item \textbf{Prompt-heavy same-budget summary}: Summary evidence showing what role each of \textbf{hotpotqa}, \textbf{multi\_news}, \textbf{qasper}, \textbf{musique}, \textbf{2wikimqa} serves as witnesses.
\item \textbf{Actual-output bridge}: Bridge evidence showing that replay-level fidelity gains are not artifacts entirely unrelated to actual generation.
\end{enumerate}

Figures are placed with the same logic. The reasoning gate frontier plot visualizes mainline reasoning evidence, and the prompt-heavy summary plot and witness map visualize witness role assignments. The actual-output bridge plot supplementarily shows the replay-to-output connection.

\subsection{Replay witness role taxonomy}

The current prompt-heavy replay witnesses should not be read as a single kind of evidence. Each witness's role should be separated as follows.

\begin{itemize}
\item \textbf{hotpotqa}: strongest same-budget witness
\item \textbf{multi\_news}: clean replay-level decode-active witness
\item \textbf{qasper}: strong crossing evidence but prefix-dominant witness
\item \textbf{musique}: weaker-but-consistent gain witness
\item \textbf{2wikimqa}: retained boundary case
\item \textbf{vcsum}: replay-level decode-active follow-up witness
\item \textbf{qmsum}, \textbf{gov\_report}: \textbf{prefix\_budget\_exhausted} boundary probes
\end{itemize}

This taxonomy is not merely descriptive; it is central to claim discipline. For example, grouping \textbf{qasper} and \textbf{multi\_news} as the same kind of decode-stage superiority evidence exaggerates the method claim. Conversely, reading \textbf{qmsum} and \textbf{gov\_report} only as failure cases misses the important lesson that the active decode regime and the prefix-budget-exhausted regime must be separated.

\subsection{What each package is supposed to prove}

The current evidence bundle falls into three major packages.

\textbf{Reasoning gate package.} This package is intended to show that CASK repeatedly demonstrates same-budget fidelity advantage in reasoning slices, with crossing against higher-budget eviction baselines appearing in some ranges. Accordingly, this package directly supports the mainline headline.

\textbf{Prompt-heavy replay package.} This package is intended to show that the two-stage policy as a whole is effective in prompt-heavy regimes and that strong same-budget gains exist. However, because this package mixes decode-active witnesses, prefix-dominant witnesses, and boundary probes, it must be interpreted only with explicit role separation.

\textbf{Actual-output bridge evaluation.} This result cluster is intended to show that replay-level fidelity gains are not entirely separated from actual generation. However, this evaluation should not be read through a single lexical-overlap score. The current output-level bridge must instead be read along three axes: \textbf{official task metric}, \textbf{lexical overlap}, and \textbf{semantic/reference similarity}. This distinction is necessary to interpret cases such as \textbf{vcsum}, where lexical score is low but reference semantics are closer, as lexical under-read rather than simple failure. This result cluster does not replace the mainline results; it functions as bridge evidence for the replay evidence.

\begin{table}[t]
\centering
\small
\begin{tabularx}{\textwidth}{YYYYYYY}
\toprule
Task & Variant & SeqRatio & SemSim & TaskMetric & SavedRatio & Read \\
\midrule
--- & --- & ---: & ---: & ---: & ---: & --- \\
qasper & TriAttention @ 512 & 0.173 & 0.678 & 11.94 & - & crossing baseline \\
qasper & CASK @ 256 & 0.238 & 0.791 & 12.77 & 90.9\% & clean three-axis crossing \\
multi\_news & TriAttention @ 384 & 0.000 & 0.452 & 0.00 & - & decode-active baseline collapse \\
multi\_news & CASK @ 384 & 0.169 & 0.952 & 15.16 & 84.1\% & strongest decode-active bridge \\
hotpotqa & TriAttention @ 256 & 1.000 & 1.000 & 27.27 & - & parity baseline \\
hotpotqa & CASK @ 256 & 1.000 & 1.000 & 27.27 & 97.6\% & output-level non-regression \\
vcsum & TriAttention @ 384 & 0.000 & 0.724 & 17.17 & - & lexical-leading boundary \\
vcsum & CASK @ 384 & 0.000 & 0.941 & 10.53 & 91.2\% & lexical-semantic split boundary \\
\bottomrule
\end{tabularx}
\end{table}

\subsection{What is appendix-only evidence}

The following materials are important but are evidence for appendix-level transparency.

\begin{itemize}
\item weighted aggregate measured counts
\item same-budget measured counts
\item budget scaling measured counts
\item raw audit matrix
\item reserve ablation wide tables
\item long-form per-task diagnostic tables
\end{itemize}

These materials serve to show what raw evidence the already-presented claims rest on, rather than to create new main-text headlines. Therefore, moving these tables to the appendix does not weaken the paper; instead, it makes the separation between the mainline narrative and the audit trail easier to read.

\subsection{Equation placement rationale}

Equations follow the same principle. The main text places only the minimum equations directly needed to understand the claims, such as the representative-set objective and metric definition formulas. In contrast, the horizon kernel, horizon-averaged mean score, RMS2 decomposition, variational horizon QP, and perturbation decomposition are placed as appendix-level assets preserving the Phase 1 mathematical archive.

This placement is not merely an editorial preference. The center of the current paper is not the general theory of score families but the transition from why-not-scorer to a selective-consolidation policy and its empirical consequences. Therefore, the formula archive is important, but it should not displace the method headline.

\subsection{Final editorial rule for the empirical package}

The final rule for organizing the evaluation structure in the current draft is as follows. The main text should contain only the minimum tables, equations, and graphs needed to build the claims. The appendix should hold the raw tables, measured counts, and mathematical archive that support those core claims. Maintaining this structure allows the main text to read as a paper while the appendix provides the necessary detailed evidence.

\bibliographystyle{plainnat}
\bibliography{references}

\CASKChecklistBlock

\end{document}